# Harmonising the Clinical Melody: Tuning Large Language Models for Hospital Course Summarisation in Clinical Coding

Bokang BI [a],[1], Leibo LIU [a], Sanja LUJIC [a], Louisa JORM [a], Oscar PEREZ-CONCHA [a]
[a] *Centre for Big Data Research in Health, The University of New South Wales, Sydney, Australia*
[1]Corresponding author: Bokang Bi, bokang.bi@student.unsw.edu.au

**Abstract:** The increasing volume and complexity of clinical documentation in Electronic Medical Records (EMR) systems pose significant challenges for clinical coders, who must mentally process and summarise vast amounts of clinical text to extract essential information needed for coding tasks. While large language models (LLMs) have been successfully applied to shorter summarisation tasks in recent years, the challenge of summarising a hospital course remains an open area for further research and development. In this study, we adapted three pre-trained LLMs (Llama 3, BioMistral, Mistral Instruct v0.1) for the hospital course summarisation task, using Quantized Low-Rank Adaptation (QLoRA) fine-tuning. We created a free-text clinical dataset from MIMIC III data by concatenating various clinical notes as the input clinical text, paired with ground truth 'Brief Hospital Course' sections extracted from the discharge summaries for model training. The fine-tuned models were evaluated using BERTScore and ROUGE metrics to assess the effectiveness of clinical domain fine-tuning. Additionally, we validated their practical utility using a novel hospital course summary assessment metric specifically tailored for clinical coding. Our findings indicate that fine-tuning pre-trained LLMs for the clinical domain can significantly enhance their performance in hospital course summarisation and suggest their potential as assistive tools for clinical coding. Future work should focus on refining data curation methods to create higher quality clinical datasets tailored for hospital course summary tasks and adapting more advanced open-source LLMs comparable to proprietary models to further advance this research.

# 1. INTRODUCTION

## 1.1 Background

Hospital course summarisation, involving generation of a concise and informative summary of the sequence of events that happen to a patient during their hospital stay, is an indispensable task to support clinical coding, as well as communication between healthcare professionals [1-4]. The hospital course summary is typically included in the patient's discharge summary, authored by a clinician who has overseen the patient's care. The implementation of Electronic Medical Records (EMRs) provides digital access to medical information, clinical assessments, diagnoses, medical imaging, and more, organised in a chronological manner. The EMR system consists of many different clinical notes, such as progress notes, radiology reports, discharge summaries, created by numerous healthcare professionals from different fields. Such extensive documentation leads to a burgeoning volume of heterogenous data contained in EMRs, thereby making navigation and comprehension challenging for both clinicians and clinical coders. Clinical coders extract relevant information from a patient's EMR and assign a unique classification code for each diagnosis and health intervention that best describe the patient's episode of care [0]. Abstracting valuable information from the clinical text occupies a significant part of their daily work of clinical coders [5-7], and only 10% of the content of EMRs is relevant to the coding task [8], making manual coding time-consuming and prone to errors [9]. Therefore, an automated method that can generate a concise hospital course summary, highlighting valuable information for clinical coders from the vast documents in EMRs, is required.

## 1.2 Related work

Automated summarisation of clinical text has been evolving over the past four decades [10]. Early works focused on extractive summarisation, in which summaries are created by selecting and listing important phrases from the original texts [7, 11]. In recent years, the development of self-attention-based Transformer models, and pre-trained Large Language Models (LLMs), have demonstrated promising capabilities in various Natural Language Processing (NLP) tasks, including, text understanding, question answering, text generation, and text summarisation [12, 13]. These advancements in NLP have shifted the summarisation focus from extractive to abstractive summarisation, creating new summaries in plain language based on the key points from the original data [14].

Clinical text summarisation has been a rising area of research, with prior studies primarily focusing on chest X-ray reports [1, 5, 15] and echocardiography [16]. In these studies, models are trained to generate the 'impression' section of the report, summarising the detailed 'findings' of the examination. While this approach produced promising results, the more complex task of hospital course summarisation using multiple sources from the EMR remains unsolved. Liang et al. used pre-transformer, convolutional neural network (CNN) models to extract key sentences from EMRs for summarising diabetes and hypertension cases [17]; Searle et al. applied transformer models, such as BERT (Bidirectional Encoder Representations from Transformers) and BART (Bidirectional and Auto-Regressive Transformers), for abstractive summarisation of EMRs by matching the brief hospital course summary in the discharge summary with corresponding descriptive sentences in the clinical notes [7]; Aali et al. assessed the effectiveness of LLMs in generating brief hospital course summaries from different sections of the discharge summary, using different LLM adaptation methods [18]. These studies highlighted the challenges of summarising clinical texts from EMRs; noting that the extensive length of the texts and the significant variability among contributors and document types make it difficult for AI models to extract the essential insights needed for clinical use [1, 5, 7, 16]. As a result, current studies focus on producing hospital course summaries, which are high-level overviews consisting of a few sentences, without extensive medical jargon or detailed information. These hospital course summaries contain scarce clinical information and are insufficient for clinical coding tasks. Therefore, a novel model adaptation benchmark, consisting of an EMR dataset that focuses on the relationship between input clinical notes and output hospital course summaries, along with a model performance validation metric for practical utility, is necessary for fine-tuning models to produce comprehensive hospital course summaries with detailed information on patients' conditions, treatments, and outcomes, suitable for healthcare professionals such as clinical coders.

This study examined how fine-tuning top-performing language models with a curated clinical dataset can help enhance their ability to summarise clinical information, replicating the way EMRs are presented in a real clinical environment. Our contributions are as follows:

- We curated a clinical free-text dataset[1] through extensive data preprocessing, cleaning and interviews with clinical coders to ensure the dataset contains as much relevant information as possible to produce summaries for clinical coding.
- We fine-tuned three pre-trained LLMs: i) Llama3, the current best-performing pre-trained LLM in the general domain; ii) Mistral Instruct v0.1, which has been successfully adopted into many different professional domains due to its high-performance fine-tuning capabilities; and iii) BioMistral, the biomedical fine-tuned version of Mistral. We enhanced its performance by extending its context window from 2048 to 4096 tokens.
- We evaluated the LLMs using automated metrics for syntactic and semantic similarity. Since current automated metrics do not reflect the accuracy and validity of clinical summaries, we designed a novel evaluation metric for clinical coding reader studies. Subsequently, we selected a random portion of these summaries to be evaluated for preference alignment with clinical coders.

## 2. METHODS

### 2.1 Data

In this study, we used the MIMIC-III dataset. This clinical dataset captures the hospital course of 53,423 admission events involving 38,597 patients in the Intensive Care Units of Beth Israel Deaconess Medical Center, USA, between 2001 and 2012 [21]). We created a clinical dataset consisting of free-text clinical notes that reflect the features of an EMR in the real world, from the de-identified free-text NOTEEVENT data from MIMIC III [19-20]. The NOTEEVENT data contains more than two million distinct clinical notes of various categories, including progress reports from nurses and physicians, study reports such as chest X-rays, discharge summaries, and more. A detailed description of the data management plan is included in Appendix A; the MIMIC III NOTEEVENT data dictionary is described in Appendix A: *Table A1*.

### 2.2 Data processing

We created a dataset that includes 33,255 EMR notes paired with their corresponding hospital course summaries. These summaries serve as the ground truth in our dataset for supervised model training. We randomly divided the dataset into 85%, 10%, and 5% for model training, validation, and test sets, respectively.

---

[1] We will release the code for data-processing pipeline, model development and evaluation at the time of publication.

A hospital course summary is intended to provide an overview of the patient's journey in the hospital, focusing on key events and clinical decisions from admission to discharge, tailored specifically for clinical coders. It emphasises coding-relevant information such as diagnoses, procedures, and treatments in a structured format that facilitates accurate coding. In the MIMIC III dataset, the hospital course summary is referred to as the 'Brief Hospital Course' (BHC) section in the discharge summary. A discharge summary is a detailed clinical document in the EMR that is written at the time of the patient's discharge from hospital, it includes other sections such as the patient's condition at discharge and any follow-up instructions. It is typically written by a clinician to support continuity of care and communication with other healthcare professionals, including primary care physicians. [6, 22].

To understand how clinical coders utilise the EMR for coding, we conducted interviews with five clinical coders. Based on their feedback and insights from previous studies on clinical text summarisation [18, 22], we identified discharge summaries as the most informative clinical notes for coding purposes. Consequently, we selected portions of the discharge summaries, particularly the BHC section, as the ground-truth for the hospital course summaries. The BHC sections were extracted using regular expressions. We tokenised the input clinical notes into tokens, the smallest units of text that the model processes, which can be as short as a single character or as long as a word or sub word unit, using LLaMA tokenizer to obtain an estimate of contextual length of the input clinical notes. For example, the sentence *'The patient was diagnosed with type 2 diabetes and prescribed metformin 500 mg daily.'* would be tokenised into the following tokens: *['The', ' patient', ' was', ' diagnosed', ' with', ' type', ' 2', ' diabetes', ' and', ' prescribed', ' met', '##formin', ' 500', ' mg', ' daily', '.']*. EMRs with either BHC sections fewer than fifty tokens or input clinical notes containing fewer than 500 tokens were excluded (3.2%), as such brief clinical reports are unlikely to provide sufficient information and are therefore deemed unsuitable to be used as reliable ground truth in our dataset.

To create paired input clinical notes for the ground-truth hospital course summaries, we extracted relevant sections from the MIMIC clinical notes, including the "assessment and plan" sections from the nursing progress reports, physician reports and other categories pertinent to generating hospital course summaries. Portions of discharge summaries are also selected to create to create the paired input clinical notes. Specifically, we used regular expressions to extract sections such as Chief Complaint, Major Procedure, History of Present Illness (HPI), Physical Exam, Discharge Diagnosis, and Discharge Disposition from discharge summaries. These sections provide valuable information for summarisation in the context of clinical coding. [6, 23-25].

Subsequently, during the dataset curation process, extensive data cleaning was performed to enhance the learning efficiency of the LLMs. The specific steps are detailed in Figure 1. We reconstructed the daily narrative of a patient's hospital stay by concatenating all the extracted texts in chronological order from the admission day to the discharge day. Sensitive information was replaced with artificial identifiers through pseudonymization to ensure privacy. Dates in the format of '[2119-01-16]' were replaced with 'Day X'. The admission date was designated as 'Day 1', and the subsequent days were numbered sequentially. Names such as 'Ms. [Known Lastname]' were replaced with 'the patient' to maintain anonymity. Additionally, the format of the EMRs was standardised by removing all excessive existing line breaks and using new line breaks to separate each daily narrative (see Table 1).

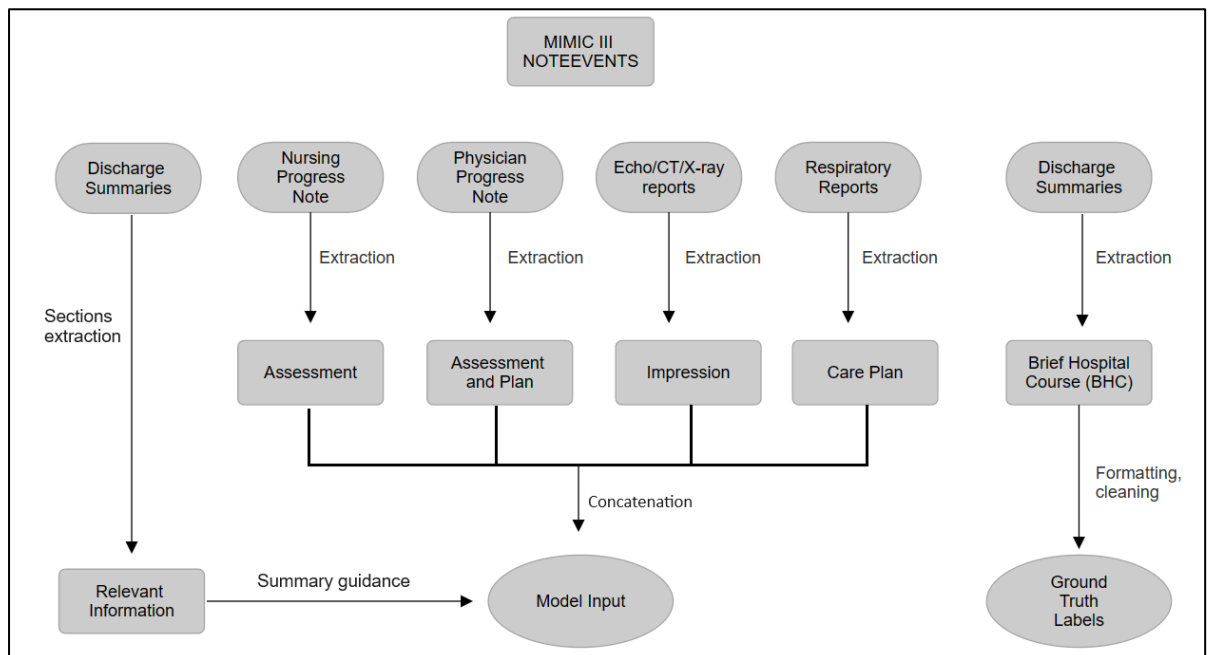

**Figure 1:** *The flow chart above illustrates the data pre-processing steps undertaken in this study. The MIMIC III NOTEEVENTS data includes various categories of clinical notes. Extracted information was concatenated with corresponding time steps <CHARTTIME> to form a single time-ordered linked file, which serves as the model input. From the discharge summaries, the Brief Hospital Course section was extracted and used as the ground-truth for model training after appropriate formatting and cleaning.*

**Table 1**. *The concatenated multi-document EMR structure in the curated clinical dataset from MIMIC III, which serves as input for model training.*

| |
|---|
| Admission day: Reason for hospitalization is {*Chief Complaints*}. History of present illness: {*History of the Present Illness*}. |
| Day 1: Extracted sections from clinical notes |
| Day 2: Extracted sections from clinical notes |
| ...... |
| Discharge day: Patient physical examination {*Physical Exams*}. Patient is diagnosed {*Discharge diagnosis*}, received {*Major Procedure*} in hospital. Patient is discharged to {*Discharge Disposition*}. |

## 2.3 Large Language models

In this study, we evaluated the effectiveness of fine-tuning a selection of the best-performing open-source LLMs as of June 2024 for clinical text summarisation within the clinical coding setting (see Table 2). These selected models, pre-trained by their respective developers, include those designed for both general-purpose use (Mistral Instruct v0.1 and Llama 3) and specialised biomedical applications (BioMistral). Detailed information about the selected models is provided in Table 2. Proprietary models such as GPT-turbo and Gemini were excluded from this study because fine-tuning these models would require uploading confidential health data to a central server [16], which violates the privacy requirements of data custodians [20].

**Table 2**: *Detailed information on the selected LLMs for this study*

| BioMistral | Mistral Instruct v0.1 | Llama 3 |
|---|---|---|

| Pre-training data | Mistral-based model further pre-trained using text data from PubMed Central | Diverse range of internet data, texts from websites, articles, books. | A new mix of publicly available online data. |
|---|---|---|---|
| Pre-training token | Detail unknown | Detail unknown | 15 trillion tokens |
| Context length | 2048 tokens | 8k Tokens | 8k Tokens |
| Model Architecture | Auto-regressive transformer model | Auto-regressive transformer model | Auto-regressive transformer model |
| Parameter size | 7 billion | 7 billion | 8 billion |
| Additional information | BioMistral is currently the best-performing open-source LLM in the biomedical domain. It was developed through continued pre-training of Mistral Instruct v0.1 using the PubMed Central dataset [27]. | Mistral Instruct v0.1 was fine-tuned from the Mistral base model using publicly available instruction datasets [26]. | Llama 3 is the current best-performing pre-trained LLM for general-purpose use. Its previous generation, Llama 2, demonstrated successful adaptation in the medical domain [1, 5, 15, 16, 18]. |

Parameter-efficient fine-tuning (PEFT), a technique that fine-tunes a small subset of model parameters, reducing computational costs while maintaining performance, has been shown to improve the performance of pre-trained LLMs for clinical text summarisation [16, 28, 29]. In this study, we used Quantized Low-Rank Adaptation (QLoRA). QLoRA delivers fine-tuning results comparable to standard Low-Rank Adaptation (LoRA) while significantly reducing the computational resources needed by using 4-bit quantisation of the LLMs' parameters [30]. Detailed hyperparameters used during the QLoRA fine-tuning are provided in the Appendix Section 2.

Instruction prompting is a technique where a model is given a role, input data, and a task description to guide its behaviour towards the desired output. We applied instruction prompting for our models to act in the roles of 'medical assistants' and to 'write a concise hospital course summary,' as described in Appendix A Table A1. This input prompt template was used for both QLoRA model fine-tuning and subsequent model inference.

### 2.4 Evaluation

The evaluation of the quality of hospital course summaries for their clinical utility is an open challenge. In this study we used four automated NLP evaluation metrics to compare the models before and after fine-tuning: the Recall-Oriented Understudy for Gisting Evaluation - Longest Common Subsequence (ROUGE-L, ROUGE-1, ROUGE-2) [31] and the Bidirectional Encoder Representations from Transformers (BERT) Score, which has demonstrated better correlations with human preferences than other existing metrics [32] (see Table 3).

Nevertheless, these metrics are limited in evaluating LLM-generated hospital course summaries because they focus on exact matches and may miss the semantic richness, coherence, and human-like understanding (see examples in Appendix C Table C1).

*Table 3*: NLP evaluation metrics

|  | ROUGE-L, ROUGE-1, ROUGE-2 [31] | BERTScore [32] |
|---|---|---|
| **Description** | Measures the matching sequences of words between reference and candidate texts. | Measures the candidate text and the reference sentences by Cosine distance similarly using BERT model context embedding. |

| Advantages | Measures recall instead of precision for better correlation with human (compared to BLEU). | Correlates with human preference (compared to BLEU/ROUGE-L). Captures semantic measurement. |
|---|---|---|
| Dis-advantages | Measures syntactical matching sequences instead of semantic matching | More computational expensive. Performance depends on underlying models |

To address the evaluation challenges and ensure that model-generated hospital course summaries meet the practical needs of clinical coders, we developed a novel evaluation metric that prioritises completeness, clinical relevance, and accuracy. Unlike previous studies that emphasised conciseness and fluency in clinical text summarisation research [1, 7, 18], our Clinical Hospital Course Summary Assessment (CHoCoSA) metric focuses on capturing the essential clinical information critical for clinical coding tasks. The CHoCoSA metric was developed based on interviews with clinical coders and findings from related study [6], which highlighted the importance of accuracy and completeness of EMR for clinical coding task. The CHoCoSA metric divides the hospital course summary into six key sub-sections—Admission Reason, History of Present Illness, Medical Assessment, Health Intervention, Diagnosis, and Discharge Information (see Table 4)—that are typically crucial for clinical coding. The CHoCoSA metric evaluates model-generated summaries against ground-truth physician-authored summaries from the MIMIC III dataset, ensuring that the summaries provide the necessary clinical details of the patients' hospital course for clinical coders. By prioritising accuracy and completeness over conciseness and fluency, the CHoCoSA metric is tailored to the practical need of clinical coders. To minimise subjectivity between reviewers and enable the potential implementation of automated CHoCoSA evaluation using AI-based techniques, we have developed a scoring system, with scores assigned as 0 (incorrect), 0.5 (incomplete or partially correct), or 1 (complete and correct). (see Appendix D). In this study, as a proof of concept and to demonstrate practicality, we applied this metric to a randomly selected set of 30 EMR inputs with paired ground-truth hospital course summaries and model-generated summaries.

*Table 4*: Examples of actual summary sub-sections in Clinical Hospital Course Summary Assessment (CHoCoSA)

| Sub-sections | Example |
|---|---|
| Admission Reason | The patient presented to OSH with SOB, found to have severe MR, admitted to hospital 118 for surgical interventions. |
| History of Present Illness | 79 y/o M with h/o HTN, HL, severe MR, and hypercholesterolaemia ... |
| Medical Assessment | Echo showed severe acute mitral regurgitation likely due to ... dilated right ventricle ... Cardiac cath showed clean coronaries and elevated both side filling pressure ... |
| Health Intervention | IABP was placed for afterload reduction and ... The patient underwent MV repair with a 27mm CE annuloplasty ring in the OR Chest tube and pacing wire, mechanical ventilation was placed post-op in CSRU ... |
| Diagnosis | Primary: Acute mitral regurgitation, post-op afib Secondary: Hypertension, hypercholesterolaemia ... |
| Discharging Information | The patient is discharged to rehab on POD #8 in stable condition |

## 3. RESULTS

### 3.1 Metric evaluation

Among the three pre-trained LLMs (referred to as PT models), BioMistral-PT consistently recorded the lowest scores across all four automated NLP evaluation metrics (BERTScore,

ROUGE-L, ROUGE-1, ROUGE-2). In contrast, Llama3-PT and Mistral Instruct v0.1-PT showed similar performance on the BERTScore, with Llama3-PT slightly outperforming Mistral Instruct v0.1-PT on the ROUGE metrics. (see Figure 2).

After fine-tuning, all three fine-tuned models (referred to as FT models) exhibited similar performance levels across the evaluation metrics. Notably, BioMistral-FT and Mistral Instruct v0.1-FT outperformed Llama3-FT on the BERTScore, ROUGE-L, and ROUGE-2 metrics. Mistral Instruct v0.1-FT obtained the highest score on ROUGE-1 compared to both BioMistral-FT and Llama3-FT. Overall, fine-tuning led to significant performance improvements in the selected models for the hospital course summarisation task, particularly in the ROUGE metrics. Additionally, the fine-tuned models displayed a convergence in performance, with the largest absolute median difference among the four metrics being less than 10% (e.g., Llama3-FT vs. Mistral Instruct v0.1-FT on the ROUGE-1 metric, 31.33 vs. 34.15).

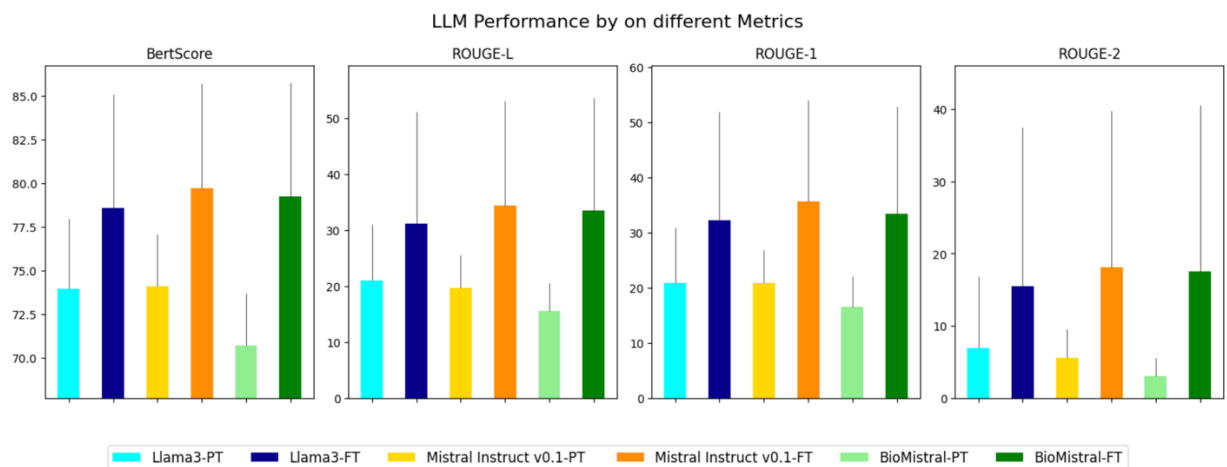

*Figure 2:* Model performance across four different automated NLP evaluation metrics (BERTScore, ROUGE-L, ROUGE-1 and ROUGE-2) with median values. The error bars on top of the bars represent the score deviation. The pre-trained models are denoted with a suffix 'PT', and fine-tuned models are denoted with a suffix 'FT'.

### 3.2 Context length performance analysis

Context length performance in LLMs refers to how effectively a model can utilise and maintain relevant information over varying lengths of context. It assesses the model's ability to generate accurate, coherent, and contextually appropriate responses when given short versus long inputs. As context length increases, performance may degrade.

In our study, three fine-tuned models: Llama3-FT (L3-FT), BioMistral-FT (BM-FT), Mistral Instruct v0.1-FT (MI-FT), and two pre-trained models: Llama3-PT (L3-PT) and Mistral Instruct v0.1-PT (MI-PT), demonstrated stable performance across both context length subgroups on the BERTScore and ROUGE metrics, with slightly better results in the longer context length subgroup (Figure 3). In contrast, the pre-trained BioMistral model (BM-PT) exhibited notably better performance in the shorter context length subgroup, particularly on the ROUGE-1 and ROUGE-L metrics. This contrasting behaviour of the BioMistral-PT model suggests that fine-tuning helped stabilised the BioMistral model's performance across longer context lengths.

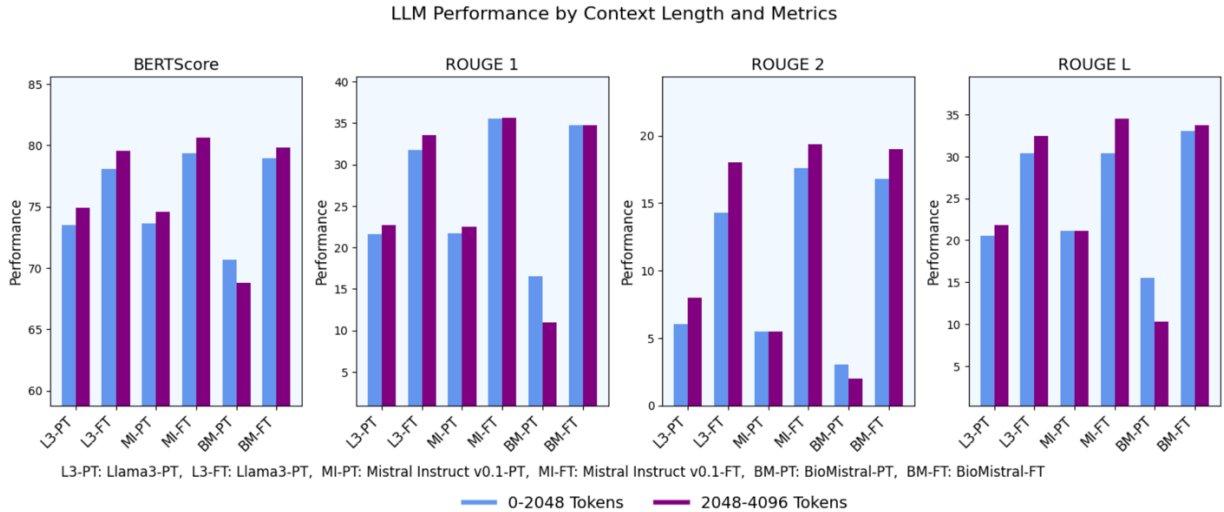

*Figure 3*: Model performance on different input token context length subgroup, across BERTScore and ROUGE metrics, before and after fine-tuning. 'L3' denotes Llama3 model, 'BM' denotes BioMistral models, 'MI' denotes the Mistral Instruct v0.1 model.

### 3.3 Clinical Hospital Course Summary Assessment

Clinical Hospital Course Summary Assessment was performed on all three fine-tuned models (Llama3-FT, Mistral Instruct-FT, and BioMistral-FT). All models demonstrated comparable results in the clinical hospital course summary assessment, with Mistral Instruct-FT performing better than Llama3-FT and BioMistral-FT (Figure 4). Analysis of hospital course summary sub-sections revealed that all three models performed well in the 'Reason for Admission' and 'History of Present Illness' sub-sections but underperformed in the 'Medical Assessment' sub-section. We observed that the Mistral Instruct-FT model performed better in summary sections containing the clinical events of the hospital course, such as 'Medical Assessment', 'Treatment', and 'Diagnosis', compared to the fine-tuned Llama3 and BioMistral models. Notably, in the 'Diagnosis' section, the fine-tuned BioMistral model underperformed considerably in contrast to the other two fine-tuned models.

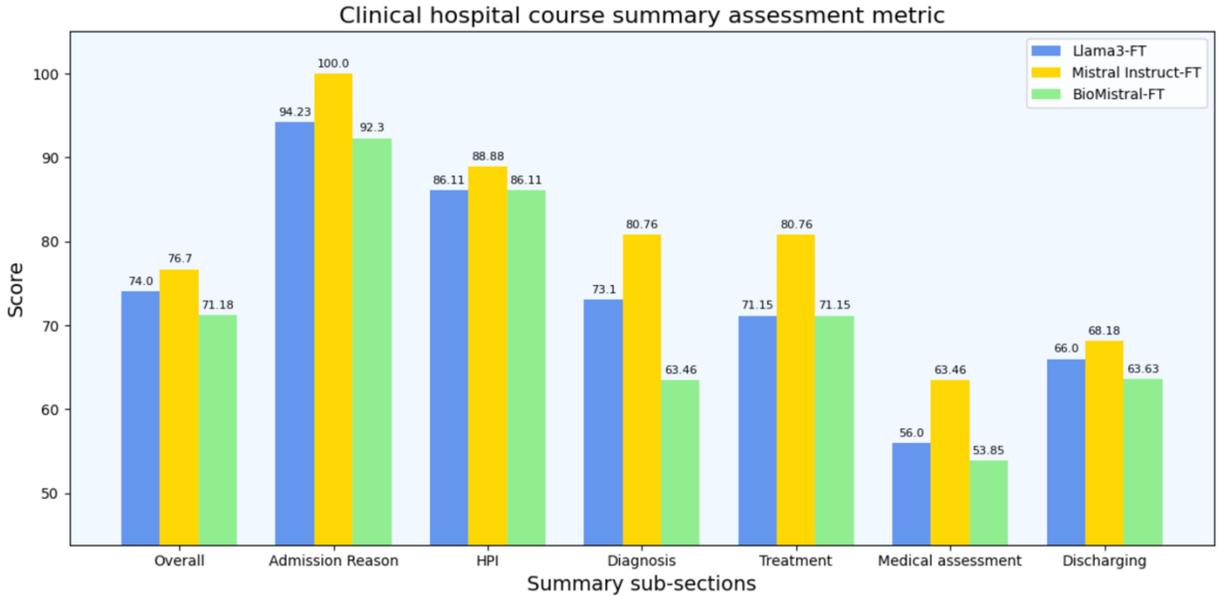

*Figure 4*: Llama3-FT, Mistral Instruct-FT and BioMistral-FT performance on the CHoCoSA metric by summary sub-sections. All three models achieved similar results in the sections describing the patient's administrative information, such as the Admission reason and History of Present Illness (HPI). Mistral Instruct-FT performed better in sections capturing the medical details of the patient's hospital course (Diagnosis, Treatment, Medical Assessment) compared to Llama3-FT and BioMistral-FT.

## 4. DISCUSSION

In this study, we investigated the domain-specific adaptation of three open-source auto-regressive LLMs for hospital course text summarisation for clinical coding, using QLoRA fine-tuning on our curated EMR dataset from the MIMIC III database. Model-generated hospital course summaries were evaluated using automated NLP evaluation metrics for syntactic similarity, and the CHoCoSA metric that assess clinical utility for clinical coding. This study found that, after sufficient training on a reasonably sized domain-specific clinical dataset, our fine-tuned LLMs showed significant improvement in their performance on downstream hospital course text summarisation tasks compared to their corresponding pre-trained models. Hospital course summaries generated by the three fine-tuned LLMs demonstrated similar median scores among automated NLP evaluation metrics, despite the initial performance disparity between the different pre-trained models. This convergence in fine-tuned models' performance suggests that fine-tuning could equip fine-tuned models with optimised strategies to meet the domain-specific requirements of the downstream task. A similar finding is illustrated in a benchmark study on generating BHC from discharge summaries using domain-adapted LLMs [18]. Among the three pre-trained (PT) models, the BioMistral-PT showed the lowest scores across all NLP evaluation metrics, likely due to its limited input context window of 2048 tokens, compared with 8192 tokens of both Llama3-PT and Mistral Instruct v0.1-PT models. Our context length study demonstrated that BioMistral-PT suffers considerable performance degradation when the input data exceeds 2048 tokens, particularly among the ROUGE's family metrics. However, fine-tuning BioMistral on clinical EMR dataset ranging from 0-4096 tokens, stabilises its performance on all four NLP metrics, and yielded a slightly better performance for the longer context sub-group, similar to Llama3 and Mistral Instruct v0.1 models. This observation is consistent with a previous study that found models fine-tuned with datasets of input token ranges from 0-4096 tend to demonstrate higher performance scores for sub-groups of longer input clinical text [18]. This observation suggest that fine-tuning could effectively extend the input context window of pre-trained LLMs.

Fine-tunning not only enhanced performance but also increased the variability in the fine-tuned models' results across all metrics, compared to the corresponding pre-trained models. This is likely due to the format variability in physician-written summaries used as ground-truth in our dataset. For instance, some physicians prefer to write summaries in a narrative paragraph form, while others opt for a bullet-point structure based on problems or diagnoses. This distinct difference in format among physician-written notes led to reduced performance scores in the automated NLP metrics when the structural format of a model-generated summary differed from that of the corresponding physician-written summary, even though both summaries might contain an equal amount of clinical information. This highlights a limitation of the current automated NLP evaluation metrics, as these metrics evaluate reference and candidate text based on the degree of syntactic overlapping (ROUGE's family metrics) or based on cosine similarity of the reference and candidate sentence embeddings (BERTScore). Consequently, both evaluation methods are sensitive to format differences between texts. Previous studies in clinical text summarisation [1, 18] have shown that automated NLP metrics often do not align with clinicians' perspectives, making human evaluation crucial in clinical applications where accuracy and relevance are critical. Thus, we evaluated the fine-tuned models using our CHoCoSA metric to assess their performance in hospital course text summarisation.

In the CHoCoSA, we observed that all three fine-tuned models (Mistral Instruct v0.1-FT, Llama3-FT, and BioMistral-FT) demonstrated similar levels of performance. In our analysis of randomly selected BHC examples, we observed that BioMistral-FT and Llama3-FT relied more on extractive summarisation compared to the Mistral Instruct v0.1-FT model, particularly in cases where the 'History of Present Illness' section from the physician admission notes is lengthy and informative. As a result, summaries generated by the BioMistral-FT and Llama3-FT had a higher lexical resemblance to the physician-written BHC in the MIMIC III, including the use of abbreviations such as POD for post operation day, and OSH for outside hospital. However, this extractive summarisation approach sometimes led BioMistral-FT and Llama3-FT to produce incomplete BHC summaries. These summaries typically included only the 'Admission' and 'History of Present Illness' sections, while omitting other critical clinical information such as 'Diagnosis' and 'Treatment'. This is reflected in the lower performance score of the BioMistral-FT and the Llama3-FT model on the 'Diagnosis' and 'Treatment' sections in the CHoCoSA metric. In contrast, Mistral Instruct v0.1-FT demonstrated the capability to generate more comprehensive BHC summaries through abstractive summarisation from the entire clinical text input.

Lacking a high-quality clinical dataset is one of the biggest challenges in hospital course summarisation. One limitation of the MIMIC III dataset is that it contains detailed daily progress notes from the ICU course but provides less clinical care data for the entire hospital course of the patient as an inpatient. This makes it particularly challenging for models to generate high-scoring summaries when the input data lacks detailed information. Additionally, this limitation contributes to score discrepancies among different summary sub-sections in the Clinical hospital course summary assessment. For example, our fine-tuned models performed generally well on the summary sections that could be derived from acute care notes (such as the Admission Reason, and History of Previous Illness sections). This information is usually found in the first few sentences of the BHC summary, as physicians usually duplicate it from the admission notes, and these admission notes are available for almost every ICU admission. Thus, this admission information is always present in the input clinical text, allowing LLMs to synthesise hospital course summaries with high level of accuracy in the admission related sub-sections.

In contrast, all three fine-tuned models performed relatively poorly on the Medical assessment and patient's discharging sections. The Medical assessment section requires detailed information about the full hospital course as this section captures information about patient's response to treatments, and patient's change in medical condition through hospital journey, and the Discharging sections details patient's condition and services during the discharge time (Appendix D *Table D1*). Therefore, some sentences in the medical assessment sections from the BHC summary ground truth do not have a paired sentences from the input clinical notes derived from the MIMIC III dataset. As a result, during the summarisation process, LLMs are more likely to hallucinate and generate information that is factually incorrect or incomplete for this sub-section, compared to the ground truth summaries written by physicians (Examples of model-generated incomplete medical assessment summaries can be found in Appendix E *Table E1*). Similar finding is also demonstrated in related study using MIMIC IV dataset for hospital course summarisation, during the reader study the reader identified some BHC in MIMIC IV seems to contain information not at all present in the provided input clinical notes [18].

One previous study addressed the limitations of the MIMIC III dataset by implementing a day-to-day summarisation approach, where content is extracted from clinical notes with matching dates to generate hospital course summaries [22]. However, this method significantly reduced the amount of available data for model training, as it required the removal of a substantial portion of text that did not correspond to the dates of specific clinical events recorded in the hospital course summaries. Another benchmark study avoids this lack of information by only using other sections in the discharge summaries as the input to synthesis the hospital course summary after removing extensive unmatched text [18]. This approach limits the model's ability to produce long and detailed hospital course summaries that meets the clinical needs for clinical coding. One possible solution to overcome the limitation in the MIMIC III dataset is to reconstruct the patient hospital course using non-text data, such as lab results, vital measurements, and pharmacy reports in combination with clinical notes as data input. The inclusion of different data would be another hyperparameter that needs to be studied in the future. Alternatively, constructing a clinical dataset from a different source than the MIMIC III database, tailored specifically for the clinical text summarisation task, with more comprehensive patient records and a focused relationship between the input clinical notes and the output hospital course summary, could be used to train LLMs that are more robust and readily implemented in a clinical setting. Therefore, we believe that at the current stage, an automated clinical hospital course summarisation tool utilising LLMs should serve as an assistive tool for healthcare professionals involved in heavy clinical information reviewing and abstracting tasks, such as clinical coders, to reduce their workload. However, their critical role in evaluation remains irreplaceable.

There were limitations in our study. One limitation is that the CHoCoSA metric is still at the proof-of-concept stage. Although the metric and scoring system were developed based on feedback and review by clinical coding teams, the scoring of the 30 selected examples was performed by a single reader with a clinical background, following the provided scoring guidelines and examples. In future research, this scoring process will be automated using LLMs fine-tuned with annotated clinical hospital course summaries based on CHoCoSA.

Finally, another limitation of this study is that we only explored a limited number of pre-trained LLMs at the 7B scale for fine-tuning. We did not include proprietary LLMs such as GPT-turbo and Gemini due to data privacy issues in clinical implementation. However, these proprietary LLMs are much larger and more capable than open-source models at the 7B scale. For future work, we aim to explore the adaptation of open-source LLMs at 70B or larger scales, as these large models have demonstrated comparable performance to proprietary LLMs.

## 5. CONCLUSION

This study found that fine-tuning pre-trained LLMs on a large, domain-specific EMR dataset, significantly improves model performance, leading to optimised solutions and performance convergence across different models in clinical settings. In CHoCoSA, we demonstrated these fine-tuned models can generate hospital course summaries encapsulating crucial information for clinical coding, including admission reasons, medical history, diagnosis, and treatment of a hospital course. The findings underscore the potential of fine-tuned models to serve as assistive tools in clinical settings, reducing the workload for healthcare professionals involved in clinical text summarisation tasks. We encourage future work to focus on crafting higher quality datasets for clinical text summarisation and exploring adaptations of models at larger scales to further improve model performance in clinical text summarisation tasks.

## ACKNOWLEDGEMENT

The experiment data is accessed via PhysioNet. This research project is funded by the Centre for Big Data Research in Health, University of New South Wales, Project ID: PS50243.

## COMPETING INTEREST

None declared.

# Appendix A Pre-fix prompting

In this appendix section, we provide the following pre-fix prompt format used for generating response from the Large Language Models (LLM), in both model fine-tuning and model inference.

*Table A1: Pre-fix prompt format*

| |
|---|
| """You are a helpful medical assistant. For the following hospital clinical notes, write a concise hospital course summary for the patient.<br><br>### Input note:<br>{row['text']}<br><br>### Response:""" |

# Appendix B Hyperparameters

*Lora configurations*

In this experiment, considering the complex structure of Electronic Medical Records (EMRs) and the need to balance model adaptation performance with computational resource costs, we loaded the pre-trained Large Language Models using a 4-bit quantisation configuration through the Bitsandbytes library. The following LoRA hyperparameters were used: LoRA rank set to 256, LoRA alpha set to 512, and LoRA dropout set to 0.1 to prevent overfitting. LoRA adaptation targeted all modules in the LLM, including q_proj, k_proj, v_proj, o_proj, gate_proj, up_proj, and down_proj, with bias set to false and random state set to 3407 for reproducibility.

*LoRA fine-tuning hyperparameters*

The training hyperparameters were as follows: batch size of 8, learning rate of 2e-5, total training steps of 12,000, warm-up ratio of 0.05, evaluation steps of 1,200, and weight decay of 0.1. Training was performed using the bf16 datatype with an AdamW 8-bit optimiser and a cosine learning scheduler. The seed was set to 3407.

*Model inference hyperparameters*

In this experiment, the hyperparameters in model inference for generating summaries on the test set, we set the temperature at 0.1, repetition penalty at 1.1, max new tokens at 1,024.

# Appendix C Traditional NLP metrics

In this Appendix, we provided some examples of clinical notes of high ROUGE-L score but are poorly aligned with expert's preferences.

*Table C1: ROUGE-L and factual correctness*

| **Physician summary:** <br> A 65-year-old male with a history of hypertension and diabetes presented with chest pain. ECG showed *myocardial infarction*. Treated with aspirin and beta-blocker. | ROUGE-L Score: |
|---|---|
| **Summary A: (Factual Incorrect)** | 0.833 |
| A 65-year-old male with hypertension and diabetes presented with chest pain. ECG showed *stroke*. Treated with aspirin and beta-blocker. | |
| **Summary B: (Factual Correct)**<br>A 65-year-old man with a history of high blood pressure and diabetes came in with chest pain. The ECG indicated a *heart attack*. He was given aspirin and a beta-blocker. | 0.565 |

## Appendix D CHoCoSA Metric

In this study, we designed a novel clinical hospital course summary assessment metric that assesses the factual correctness of clinical information in model-generated hospital course summaries. This assessment is done by comparing the summaries against both the input clinical text and physician-written summaries. Each summary is subdivided into six parts: Admission reason, HPI (History of Present Illness), Medical assessment, Health intervention, Diagnosis, and Discharging Information. These sections capture different aspects of the hospital course summary, as described in Table D1: Clinical information in summary sub-sections.

*Table D1: Clinical information in summary sub-sections*

| Sub-sections | Clinical information |
|---|---|
| Admission Reason | This section captures the patient's admission information, such as symptoms, complaints, initial diagnosis by the admitting physician, and the type of service: elective procedure or emergency. |
| History of Present Illness | This section captures the patient's demographic information, such as age and sex, as well as a brief past medical history, including co-morbidities from previous hospital admissions.<br>Since every patient in the MIMIC III dataset was admitted through the ICU, this section also includes a summary of the ICU course, such as the patient's treatments in the ICU and their response to these treatments. |
| Medical assessment | This section focuses on diagnostic interventions, such as X-rays, echocardiography, and cardiac catheterisation. These procedures are usually performed to confirm the patient's diagnosis, for example, using echocardiography to confirm aortic valve stenosis. The patient's response to these interventions is also included, such as bleeding as an adverse effect. |
| Health Intervention | This section details the treatments the patient received during the hospital course, including both procedures (such as surgery) and medications (such as diuretics). The indication for each treatment is also included, for example, aortic valve replacement to treat aortic valve stenosis, and diuretics to reduce pre-operative weight. |
| Diagnosis | This section is crucial for clinical coding as it contains the patient's primary diagnosis during the hospital course. It also includes secondary diagnoses resulting from complications of the hospital course, such as medication side effects due to reduced liver function and chronic hepatic failure. |
| Discharging Information | This section describes the patient's discharge, including their condition at the time of discharge and the discharge destination, such as a rehabilitation centre, external hospital, nursing home, home, or if the patient passed away during the hospital course. |

As described in the method above, during evaluation, each section in the hospital course summary is assigned a score of 0 (incorrect), 0.5 (incomplete), or 1 (complete and correct) against the input clinical text and physician summary. Table D2 below describes examples of summary parts at different levels of accuracy from the clinical accuracy analysis.

*Table D2: Clinical hospital course summary assessment metric examples*

| Sub-sections | Examples from clinical accuracy analysis |
| --- | --- |
| Admission reason | *Physician summary (MIMIC III):*<br><br>Pt was admitted through the emergency department for subarachnoid hemorrhage.<br><br>*Complete and correct summary – model generated (1/1):*<br><br>Pt was admitted to the ICU for close monitoring after being transferred from an OSH with a sub arachnoid hemorrhage.<br><br>*Incomplete summary – model generated (0.5/1):*<br><br>Pt admitted to ICU for close monitoring. (No mention of admission reason) |
| History of Present Illness | *Physician summary:*<br><br>The patient was a 45 yo female with history of Hepatitis C cirrhosis, alcohol abuse, and grade II esophageal varices<br><br>*Complete and correct summary – model generated (1/1):*<br><br>45 yo F with HCV cirrhosis, EtOH abuse, hx of esophageal varices<br><br>*Incomplete and incorrect summary – model generated (0/1):*<br><br>The patient was admitted to the MICU after becoming hypotensive... (No mention of relevant past medical history) |
| Medical Assessment | *Physician summary:*<br><br>An upper endoscopy showed a large ulcer with a visible bleeding vessel located in the stomach antrum... A second upper endoscopy was performed showing grade II esophageal varices with evidence of recent bleeding but no active bleeding...<br><br>*Complete but contains incorrect information – model generated (0.5/1):*<br><br>EGD showed only grade I varices and an esophageal ulcer... A repeat EGD was done which showed no evidence of active bleeding. (Model summary reported grade I varices instead of grade II in the physician summary) |
| Health Intervention | *Physician summary:*<br><br>The patient underwent an aortic valve replacement ... transferred to the CVICU for invasive monitoring ... weaned from sedation awoke neurologically intact and extubated. Beta blockers and diuretics were started and he was diuresed towards his pre-op weight. Chest tubes and epicardial pacing wires were removed ...<br><br>*Complete and correct summary – model generated (1/1):*<br><br>The patient underwent an aortic valve replacement ... ... was transferred to the CVICU intubated and sedated. ......awoke neurologically intact and was weaned from the ventilator and extubated without difficulty. All lines and drains were discontinued ... Beta blockers were initiated and titrated for maximum effect. ... diuresed towards pre-op weight<br><br>*Incomplete summary – model generated (0.5/1):*<br><br>The patient underwent an aortic valve replacement ... intubated and sedated. ......awoke neurologically. <u>(contains the intervention, but missing patient's response to the procedure, and missing medication treatment following the procedure)</u> |
| Diagnosis | *Physician summary:*<br><br>Aspiration pneumonia... hypotension... A-fib... CHF... Alzheimer... Thrombocytosis... Anemia... Lethargy<br><br>*Incomplete summary – model generated (0.5/1):*<br><br>Respiratory distress... hypotension... a-fib, Alzheimer disease...<br><br><u>(Covers all the primary diagnosism, and some secondary diagnosis, but missing additional relevant secondary diagnosis which results in the complication of hospital course: CHF, Thrombocytosis)</u> |
| Discharging Information | *Physician summary:*<br><br>Cleared for discharge toin Hospital... rehab on post-op day 8<br><br>*Complete and correct summary – model generated (1/1):*<br><br>ready for discharge to rehab on POD 8. |

# Appendix E Medical assessment section comparison

In Table E1 below, we demonstrate one set of summaries generated by different fine-tuned models and the ground truth summary for the same clinical EMR input. The underlined text in the ground truth summary is the 'medical assessment' information that is not captured in the clinical notes in the MIMIC dataset. The missing 'medical assessment' information includes the diagnostic procedure chest X-ray (CXR) for the confirmation of respiratory tract infection and the laboratory studies, such as full blood tests and urine tests, to confirm the patient's anaemic condition.

*Table E1: Example model generated summaries of missing 'medical assessment' section*

| Summary source | Ground Truth | Llama 3 | Mistral Instruct v0.1 | BioMistral |
|---|---|---|---|---|
| **Summary** | Pt admitted for aortic valve replacement... transferred to CSRU... weaned from sedation and Neo-synephrine for BP support ... started on beta-blockers, aspirin, diuresis to pre-op weight. <u>CXR revealed no pneumothorax but possible left pneumonia with biliteral effusion... Pt was anemic with HCT of 22, platelet count trending down, HIT screen was negative... developed fever of 103, WBC was normal.. Urine and blood culture negative..</u> Epicardial pacing wires removed... discharged in stable condition | Pt admitted for aortic valve replacement... transferred to CSRU... weaned from mechanical ventilation and sedation, Neo Synephrine... Chest tubes removed... started on beta blockers and diuresed towards pre-op weight... transferred to floor and epicardial pacing wires removed ... discharged home with VNA services | Mr... underwent Aortic Valve Replacement ... transferred to CSRU... weaned from sedation... started on beta blockers and diuresed towards pre-op weight... transferred to floor... chest tubes were removed ... epicardial pacing wires were removed ... discharged home with VNA services | Pt admitted for aortic valve replacement... transferred to CVICU... weaned from pressors, Neosynephrine... started on beta blockers and diuresed towards pre-op weight... transferred to floor and epicardial pacing wires removed ... discharged home with VNA services |